\documentclass{article} 
\usepackage{techreport_onecol}

\usepackage{times}
\usepackage{microtype}
\usepackage{epsfig}
\usepackage{graphicx}
\usepackage{amsmath}
\usepackage{amssymb}

\usepackage{color}
\usepackage{ccicons}    
\usepackage{txfonts}
\usepackage{textcomp}

\usepackage{url}
\usepackage{multirow}
\usepackage{booktabs}
\usepackage[ruled]{algorithm2e}
\usepackage{mathtools}
\usepackage{tabularx}
\usepackage{varwidth}
\usepackage{stmaryrd}
\usepackage{balance}

\usepackage[capbesideposition=inside, facing=yes,capbesidesep=quad]{floatrow}

\newcommand{\mbe}{\mathbf{e}}
\newcommand{\mbf}{\mathbf{f}}
\newcommand{\mbm}{\mathbf{m}}
\newcommand{\mbr}{\mathbf{r}}
\newcommand{\mby}{\mathbf{y}}
\newcommand{\calA}{\mathcal{A}}
\newcommand{\calB}{\mathcal{B}}
\newcommand{\calI}{\mathcal{I}}
\newcommand{\calX}{\mathcal{X}}
\newcommand{\calY}{\mathcal{Y}}
\newcommand{\R}{\mathbb{R}}

\makeatletter
\DeclareRobustCommand\onedot{\futurelet\@let@token\@onedot}
\def\@onedot{\ifx\@let@token.\else.\null\fi\xspace}
\def\eg{{e.g}\onedot} 
\def\ie{{i.e}\onedot}

\def\etal{\emph{et al}\onedot}
\makeatother

\usepackage[pagebackref=true,breaklinks=true,letterpaper=true,colorlinks,bookmarks=false]{hyperref}
\begin{document}

\title{Criteria Sliders: Learning Continuous\\Database Criteria via Interactive Ranking}
\author{James~Tompkin,$^{1*}$ Kwang~In~Kim,$^{2*}$ Hanspeter Pfister,$^3$ and Christian~Theobalt$^4$\\
$^1$Brown University
$^2$University of Bath \\
$^3$Harvard University 
$^4$Max Planck Institute for Informatics 
}
\date{}

\maketitle

\begin{abstract}\noindent
Large databases are often organized by hand-labeled metadata, \textit{or criteria}, which are expensive to collect. We can use unsupervised learning to model database variation, but these models are often high dimensional, complex to parameterize, or require expert knowledge. We learn low-dimensional continuous criteria via \emph{interactive ranking}, so that the novice user need only describe the relative ordering of examples. This is formed as semi-supervised label propagation in which we maximize the information gained from a limited number of examples. Further, we actively suggest data points to the user to rank in a more informative way than existing work. Our efficient approach allows users to interactively organize thousands of data points along 1D and 2D continuous sliders. We experiment with datasets of imagery and geometry to demonstrate that our tool is useful for quickly assessing and organizing the content of large databases.
\end{abstract}

\section{Introduction}
\begin{figure}[t]
    \centering
		\includegraphics[width=\linewidth]{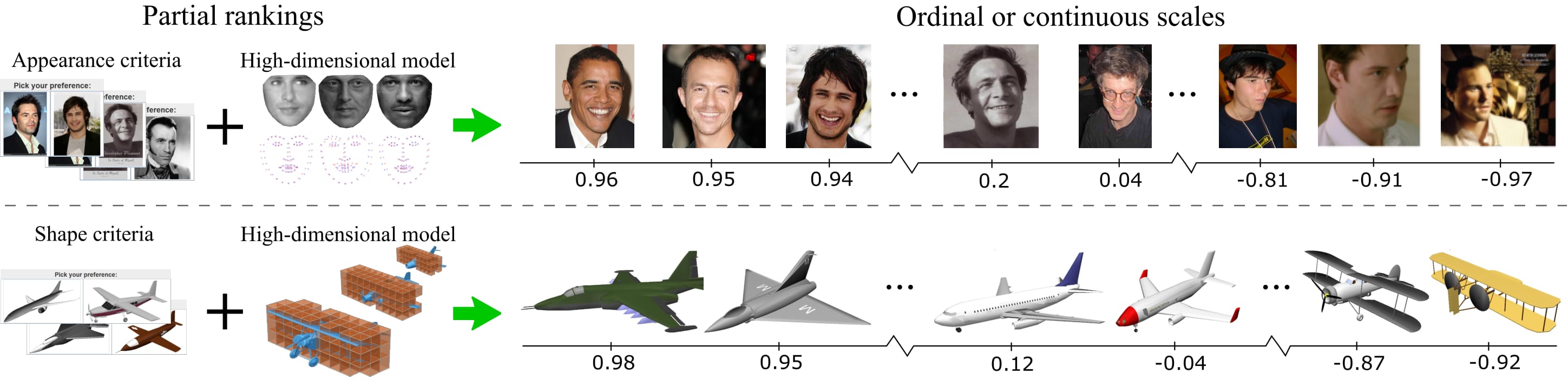}
    \caption{Users generate continuous criteria \emph{(right)} interactively by ranking examples \emph{(left)}. \emph{Top:} A database of face images is ordered on a `smiling expression' 1D parameterization. \emph{Bottom:} A database of aircraft geometry is ordered by the user criteria of `sleekness'.}
    \label{fig:teaser}
\end{figure}

Database organization often requires laborious metadata annotation. For visual data, computer vision helps automatically model the database with high-dimensional 2D image and 3D shape features. From these, we parameterize \emph{criteria} for easy organization by embedding the high-dimensional representations into a low-dimensional space, \eg, for face expression, ascribing `amount of smile' from the high-dimensional features. This task is complicated, because most desired criteria do not map to individual features, and instead map to complicated paths lying on a manifold in the high-dimensional feature space. As such, this parameterization is accomplished either by an expert, or by supervised learning from labeled examples \cite{ParGra11,ChenBennett2013}.

It is hard to survey a visual database of even a few thousand items, and even an expert would need time to assess the database for parameterization. This is especially time consuming for new databases, \eg, scraping the Internet and recovering unknown contents. Even after parameterization, organization tools are still restricted to only the expert-defined criteria, with no easy way for users to define new criteria for their interests, especially if they are abstract or cross typical boundaries. The ability to \emph{interactively} parameterize a database is needed: to quickly describe database variation without prior knowledge or labels; to intuitively discover low-dimensional criteria from high-dimensional models.

We present an interactive system to generate continuous criteria from high-dimensional models. Interactive labeling requires efficient computation, and ideally extracts the most information from the fewest user-provided labels. Thus, we adapt a state-of-the-art semi-supervised learning algorithm for interactive use, to exploit the rich structure of unlabeled data points. To best exploit limited user interaction, our system actively suggests data points to label such that the information gain is maximized. Further, we propose a new sparse large-scale active learning strategy that enables interactive label suggestion for large datasets.

To simplify parameterization, we ask the user to \emph{rank} examples. This removes much of the burden when describing potentially abstract criteria on continuous scales. For instance, deciding that item 1 is more than, equal to, or less than item 2 is easier than quantifying that item 1 is 0.2 criteria units away from item 2. From this ranking, our semi-supervised approach generates continuous criteria, which become \emph{sliders} in our user interface. For 2D criteria, users provide example embeddings directly such that the underlying relative locations are automatically meterized.

Our problem is not conventional ranking for data retrieval applications, where the goal is to classify all data instances that \emph{match} the given query. Instead, we wish to regress \emph{all} database items into continuous 1D and 2D criteria. We present two motivating scenarios: 1) For criteria which have no well-defined answer, our system helps users define their opinion. 2) For new visual databases with no metadata, where the alternative may be to laboriously hand collect labels, our system helps to quickly assess and organize the variation within the database.

We show the generality of our approach across databases of images of paintings and faces, and of geometries of human bodies and man-made objects (Fig.~\ref{fig:teaser}). We contribute:
\begin{enumerate} \itemsep0pt
\item The scenario of interactively defining continuous criteria from high-dimensional models, with a prototype implementation (please see our supplemental video).
\item A maximally-informative efficient semi-supervised active label suggestion algorithm.
\end{enumerate}

\section{Related work}
\paragraph{Re-ranking.}
This related problem takes the results of context- or text-based search and refines the query and/or retrieved result with user interaction. Personalized faceted search~\cite{KorZhaLiu08} exploits relevant meta-data and suggests new keywords to refine the current search. User behavior is modeled probabilistically and tuned to maximize the expected \emph{utility} of the facet. A rich literature shows the importance of re-ranking in search \cite{KorZhaLiu08}; however, existing algorithms in this context focus on \emph{maximizing search efficiency} rather than organizing databases along criteria.

Jain and Varma~\cite{JaiVar11} assume click behaviour relates to the interest query, and use a click count model to predict relevant rankings. Zha~\etal~\cite{ZhaYanMei10} proposed a similar approach for visual query suggestion. The COPE system~\cite{BalHalJos12} interactively refines search queries by users stating whether the results match their information need, which then weights image features for future searches. These approaches re-rank imagery based on user confirmation, and so they typically focus on the discrete problem of whether the retrieved results is a match~\cite{KorZhaLiu08}. Our goal is to learn criteria outright from sparse user labels and a high-dimensional model.

\paragraph{Rank learning.}
Existing rank learning algorithms typically use supervised learning from labeled data points (\eg, rank support vector machines; RankSVMs), whereas in our interactive setting, the user starts with no labels. As such, we exploit information in the unlabeled data with semi-supervised learning~\cite{ChaSchZie06,Zhu08}, which has been used in rank aggregation~\cite{CheWanSon08} though not in our rank propagation case. Semi-supervised learning approaches often rely heavily on graph Laplacian regularization, which is not suitable for learning on high-dimensional manifolds~\cite{NadSreZho09,KimTompkin2015}. As our data lie in high-dimensional spaces, our approach overcomes this limitation. Szummer and Yilmaz~\cite{SzuYil2011} apply the graph Laplacian to the orthogonal problem of learning a preference criteria, and our learning approach could improve this application.

For data retrieval, Parikh and Grauman \cite{ParGra11} learn discrete ranking functions via RankSVM from existing user labels. This restricts exploration to known criteria, whereas we discover criteria interactively. Murray~\etal~\cite{MurMarPer12} presented a database for visual analysis that is characterized by abstract `aesthetic' features, while Caicedo~\etal~\cite{CaiKapKan11} exploited user preference for image enhancement. Reinert~\etal~\cite{ReiRitSei13} use interaction to visually arrange a small image database into an aesthetic overview, which is orthogonal to the efficient exploration that we pursue.

Our interactive criteria definition on high-dimensional data does not compare directly to existing supervised criteria learning systems. CueFlick \cite{FogTanKap08,AmeFogKap11} learns on binary labels, and WhittleSearch \cite{ParGra11} attempts to re-rank data along existing criteria rather than generate criteria from scratch. We improve upon WhittleSearch's underlying RankSVM techniques when adapted to our scenario (Sec.~\ref{sec:experiments}).

\paragraph{Active learning.} Chen~\etal~\cite{ChenBennett2013} apply active learning to remove inconsistency from existing crowdsourced labels. Their non-interactive approach is approximate in information gain, while our interactive approach is exact given model assumptions. Shen and Lin~\cite{ShenLin2013} essentially use RankSVM for bipartite ranking, with active learning based on single point and pair closeness. This is very similar to baseline predictive variance, which may accidentally pick uninformative outliers. Our new measure is unbiased by outliers. Our active learning approach is complementary to human-in-the-loop active learning approaches, \eg, Branson~\etal~\cite{BraHorWah14}. Their task is to select \emph{features} for a given data point which minimize class conditional distribution uncertainty (\eg, 20 questions game). Our problem is to suggest \emph{data points} to label. Fogarty~\etal~\cite{FogTanKap08} learn image retrieval criteria from binary labels, \eg, outdoor vs. indoor, by iterative refinement of distance measures between data points. Their active label suggestion was extended by Amershi~\etal~\cite{AmeFogKap11} by adopting a Gaussian process model on distance measures. We ask users to provide rank labels, and increase performance over Amershi~\etal~(Sec.~\ref{sec:expactivelearning}). 

\paragraph{Concept embedding.} Our approach can be interpreted as using interaction to embed data into a high-level concept space. Existing work in this area focuses on category- or cluster-level supervision. Wilber~\etal~\cite{WilKwaKri15} receive triplet constants from users ($(i,j,k)$: object $i$ should be closer to object $j$ than it is to $k$) to learn pair-wise similarity kernels that are used in $t$-SNE-type embedding~\cite{MaaHin08}. We ask users to provide rank labels and emphasize continuous parameterization.

\section{Semi-supervised criteria learning}
\label{sec:semisupervisedlearning}
From a large database of images or geometry, we compute offline a high-dimensional vector of 2D and 3D features (Sec.~\ref{s:databases}). We assume that some combination of these are sufficient to describe the desired user criteria, with our problem being to learn the criteria from a very small number of samples. We use semi-supervised learning to compensate for the lack of labeled data points by exploiting the \emph{rich structural information} contained within unlabeled data points. Given this preprocess, the user produces a ranking (with equality) for a subset of the dataset items with our interface (Sec.~\ref{s:prototype}, and our supplemental video). The labels for these ranked examples are assigned an internal continuous representation ranging from -1.0 to 1.0.

Formally, for the set of data points $\calX = \{X_1,\ldots,X_u\}$, plus the corresponding \emph{labels} for the first $l$ data points, $\calY =\{Y_1,\ldots,Y_l\}\subset \R$, where $l\ll u$, the goal of semi-supervised learning is to infer or propagate to the labels of the remaining $u-l$ data points in $\calX$. We adopt the standard energy minimization approach~\cite{ZhoBouLal04,ZhuLafGha03,ChaSchZie06}:
\begin{align}
\label{e:energyreform}
E(\mbf) =(\mbf-\mby)^\top L (\mbf-\mby)+\lambda \mbf^\top H \mbf,
\end{align}
where $L$ is a diagonal label indicator matrix and $\mby$ is a vector of continuous label values: $L_{[i,i]}=1$ and $\mby_i=Y_i$ if $i$-th data point is labeled, and $L_{[i,i]}=0$ and $\mby_i=0$, otherwise. $H$ is the \emph{regularization matrix} which quantifies how to \emph{smooth} the propagated labels $\mbf$ within their local context and the regularization hyper-parameter $\lambda$ balances between the smoothness of $\mbf$ and the deviation from the labels $\mby$. In general, for a symmetric non-negative definite matrix $H$, the energy functional $E$ is convex with respect to $\mbf$. The solution $\mbf^*$ is then explicitly given as:
\begin{align}
\mbf^* =(L+\lambda H)^{-1}L \mby.
\end{align}

One of the best established regularizers $H$ for semi-supervised learning is the graph Laplacian matrix $S$~\cite{Lux07,ZhoBouLal04,ZhuLafGha03} which measures pair-wise deviations of $\{\mbf_i\}$ weighted by the similarity of the corresponding inputs. With predominant use of the graph Laplacian regularizer, semi-supervised learning has become established in classification and clustering tasks where the output is a \emph{discrete} set of labels. However, estimating \emph{continuous} outputs is still an area of active research due to degenerate behavior of the graph Laplacian regularizer: in high-dimensional spaces, minimizing $E$ with $H=S$ tends to produce an output $\mbf^*$ where most elements are close to zero~\cite{NadSreZho09}. This is fine for classification as output magnitudes are irrelevant, \eg, to classify data points into \emph{positive} and \emph{negative} classes, $1$ and $1^{-10}$ are both positive; however, having close to zero outputs is meaningless for continuous outputs, as very little regularization occurs.

This problem has been approached with the \emph{local Gaussian} (LG) regularizer \cite{KimTompkin2015}, which explicitly addresses this degeneracy and reduces the computational complexity required to remove it. However, since the LG regularizer was developed for general batch semi-supervised learning, \ie, all labeled data points are provided before the learning process, it cannot be directly applied to interactive ranking. We adapt the LG regularizer to our setting, and demonstrate that it is more suitable than the graph Laplacian regularizer for cases with few labels (Sec.~\ref{s:databases}).

\subsection{Interactive ranking and active label suggestion}
\label{sec:activelabelsuggestion}
In interactive ranking, the user iteratively provides training labels until they are happy with the learned criteria. Na\"{i}vely using the LG regularizer is computationally expensive for large datasets since, at each iteration, obtaining the propagated labels $\mbf$ requires minimizing $E$ in Eq.~\ref{e:energyreform}, which is order $O(u^3)$ complexity and requires solving a linear system of size $u\times u$. Further, we wish to aid the user by suggesting data points to label. At iteration $t$, we wish to estimate criteria uncertainty per data point from existing labels, and present the user with more informative samples to label at time $t+1$. 

Following the analogy between regularized empirical risk minimization and maximum a posteriori (MAP) estimation~\cite{RasWill06}, and by adopting Bayesian optimization~\cite{SnoLarAda12}, we reformulate $-1\times E$ (Eq.~\ref{e:energyreform}) as (the logarithm of) a product of the prior $p(\mbf)$ and a Gaussian noise model $p(\mby|\mbf)$. This leads us to assess the minimizer of $E$ as the mean of the predictive distribution (the \emph{posterior}):
\begin{align}
-\log p(\mbf|\mby) = (\mbm-\mbf)^\top C^{-1}(\mbm-\mbf) + Z,
\label{e:bayesian}
\end{align}
with mean $\mbm=C L \mby$, covariance matrix $C=(L+\lambda H)^{-1}$, and the normalization constant $Z$. This perspective informs a \emph{predictive uncertainty} for each data point: The $i$-th diagonal component $C_{ii}$ of the covariance matrix $C$ contains information on the uncertainty of the prediction on label $\mbf_i$, which is typically low when $X_i$ is labeled and is high otherwise. 
Active label suggestion presents results which minimize uncertainty. 

This Bayesian reformulation is mathematically rigorous: By the construction of $H$~\cite{KimTompkin2015}, $C^{-1}$ is positive definite and so $\mbm$ and $C$ are valid mean vector and covariance matrix. This shapes the posterior $p(\mbf|\mby)$ as a sample from an underlying Gaussian process that factorizes to a prior on $\mbf$ and the corresponding likelihood $p(\mbf|\mby)$. This re-interpretation enables us to exploit well-established Bayesian uncertainty modeling. In particular, the variance of the predicted distribution $p(\mbf|\mby)$ indicates how \emph{uncertain} we are about each prediction we make on data points $X\in\calX$. 

One simple and well-established strategy to exploit these modeled uncertainties for active label selection is to predict at each iteration $t$, and choose point $X_i$ with the largest uncertainty. However, Figure~\ref{f:als} shows that naively choosing data points with maximum uncertainty leads to poor performance with a higher error rate than random selection, as isolated outlier data points---which are not broadly informative---receive high variances and are chosen. 

Instead, we construct the candidate data points that minimize the predictive variance over the \emph{entire} set of data points. At each time step $t$, we choose data points with the highest average \emph{information gain} $\calI$, defined as:
\begin{align}
\calI(X_i) = \sum_{j=1,\ldots,u} [C(t-1)-C(t)^i]_{jj}.
\label{e:infogain}
\end{align}
where $\text{diag}[C]$ is a vector containing the diagonal components of the matrix $C$. The matrix $C(t)^i$ is constructed by adding 1 to the $i$-th diagonal element of $C(t-1)^{-1}$ and inverting it (\ie, $i$-th data point is regarded as labeled; see Eq.~\ref{e:energyreform}). As suggested by the form of the matrix $C$, this score can be calculated without knowing the corresponding label value $Y_i$. This is due to the Gaussian noise model; in general, different noise models lead to the predictive variance ($C$) depending on the labels.

Na\"{i}vely estimating the information gain for all data points requires quadratic computational complexity: One has to estimate the minimizer of $E(\mbf)$ (Eq.~\ref{e:energyreform}), which is $O(u^3)$ for each data point. However, in our iterative label suggestion scenario, $\calI$ can be efficiently computed in linear time: Assuming that $C(t-1)$ is given from the previous step, calculating $\text{diag}[C(t)^i]$ does not require inverting the matrix $L(t)+\lambda H$: Using \emph{Sherman--Morrison--Woodbury} formula, $C(t)^i$ can be efficiently calculated from $C(t-1)$:
\begin{align}
\text{diag}[C(t)^{i}]=\text{diag}[C(t-1)]-\frac{\text{squ}[C(t-1)_{[:,i]}]}{1+\text{diag}[C(t-1)]_{i}},
\label{e:rankoneupdate}
\end{align}
where $A_{[:,j]}$ denotes a vector formed from the $j$-th column of the matrix $A$, $\text{diag}[A]$ constructs a vector from the diagonal components of matrix $A$, and $\text{squ}[B]$ is a vector obtained by taking element-wise squares of the vector $B$. Accordingly, after explicitly calculating $C(0)$,\footnote{In practice, we limit $C(0)$ to 2,000 labels for faster interactivity.} subsequent updates in $C(t)$ and $C(t)^i$ can be performed efficiently for each iteration $t$.

\paragraph{Large-scale extension.}
The rank-one update model (Eqs.~\ref{e:infogain} and \ref{e:rankoneupdate}) relies on explicitly calculating the covariance matrix $C=(L+\lambda H)^{-1}$. Even though the original regularization matrix $L+\lambda H$ is sparse, its inverse $C$ is generally dense. Therefore, this strategy is not directly applicable for large-scale problems where calculating and storing the inverse of $L+\lambda H$ is infeasible. For this case, we adopt sparse eigen-decomposition-based approximation of $L+\lambda H$.

Assume that $L+\lambda H$ is eigen-decomposed:
\begin{align}
C^{-1}&=L+\lambda H= E^F\Lambda^F {E^F}^\top\nonumber \Leftrightarrow C= E^F{\Lambda^F}^{-1} {E^F}^\top=\sum_{i=1}^u \frac{1}{\lambda_i}\mbe_i\mbe_i^\top,\nonumber
\label{e:approx}
\end{align}
where $u\times u$-sized matrix $E^F$ stores eigenvectors ($\{\mbe_i\}_{i=1}^u$) column-wise, and $\Lambda^F$ is a diagonal matrix of the corresponding eigenvalues ($\{\lambda_i\}_{i=1}^u$). We assume that the eigenvalues and the corresponding eigenvectors are arranged in increasing eigenvalues. Given the prescribed rank parameter $r$, our approximate covariance matrix $\overline{C}$ is given as the best possible $L^2$-approximation of rank $r$ to $C$:
\begin{align}
C\approx\overline{C}= \sum_{i=1}^r \frac{1}{\lambda_i}\mbe_i\mbe_i^\top=E \Lambda^{-1} E^\top,
\end{align}
$E$ is a $u\times r$ submatrix of $E^F$ containing the first $r$ eigenvectors, with $\Lambda$ as the corresponding $r\times r$ submatrix of $\Lambda^F$.

Now suppose that we add a label of the $i$-th data point at time $1$. Applying the Sherman--Morrison--Woodbury formula to $\overline{C}$, we can represent the update equation:
\begin{align}
C(1)^i &=E \Lambda^{-1}E^\top - \frac{E \Lambda^{-1} E_{[i,:]}^\top E_{[i,:]} \Lambda^{-1}(t-1) E^\top  }{1+H^{-1}_{ii}}\nonumber\\
&=E \left(\Lambda^{-1}  - \frac{\Lambda^{-1} E_{[i,:]}^\top E_{[i,:]} \Lambda^{-1} }{1+H^{-1}_{ii}}\right)E^\top.\nonumber
\end{align}
The corresponding information gain is given as:
\begin{align}
\text{diag}[C-C^i]_k &= E_{[k,:]} \left(\frac{\Lambda^{-1} E_{[i,:]}^\top E_{[i,:]} \Lambda^{-1}}{1+H^{-1}_{ii}}\right)E_{[k,:]}^\top \nonumber\\
&= \left(\frac{1}{1+H^{-1}_{ii}}\right) \left(E_{[k,:]}\Lambda^{-1} E_{[i,:]}^\top E_{[i,:]} \Lambda^{-1}E_{[k,:]}^\top\right).
\label{e:updateequ}
\end{align}
At step $t$, adding a label to the $i(t)$-th data point leads the a new covariance estimate:
\begin{align}
[EF_iE^\top]_{k,k} &= E_{[k,:]}\Lambda^{-1}E_{[k,:]}^\top+\sum_{i=1,\ldots,t} E_{[k,:]}\mbr_i \mbr_i^\top E_{[k,:]}^\top,\nonumber
\label{e:eigenupdate}
\end{align}
where:
\begin{align}
\mbr_i= \left(\frac{1}{1+\overline{C}^{-1}_{ii}}\right)E_{[i,:]} \Lambda^{-1}.
\end{align}

\paragraph{Amershai~\etal comparison.}
One of the first algorithms to propose active learning in end-user interaction application is Amershi~\etal's approach~\cite{AmeFogKap11}. Their algorithm was originally developed for active labeling in metric learning contexts, and so it is not directly applicable to our problem. However, the re-interpretation of their label selection criterion enables transfer to our setting. Their strategy is based on a Gaussian process (GP) model with Gaussian noise: For a given set of labeled data points $S$ and the unlabeled data points $U$, and for the $i$-th unlabeled point, the corresponding score $f(i)$ is defined as:
\begin{align}
f=\frac{Var(i|S)}{Var(i|U)},
\end{align}
where $Var(i|S)$ is the variance predicted for $i$-th data point based on the GP model trained based on $S$, and $Var(i|U)$ is the variance predicted based on the GP model trained on $U$, assuming that $U$ are all labeled. Details of GP prediction can be found in~\cite{RasWill06}. This interpretation enables us to apply the same principle to our setting: for the $i$-th data point, the corresponding score $\calI'(X_i)$ is:
\begin{align}
\calI'(X_i) = \frac{\text{diag}[C(t-1)]_i}{\text{diag}[U(t-1)]_i},
\end{align}
where $U(t-1)$ is the variance predicted from all unlabeled data points. Maximizing only the predictive uncertainty $\text{diag}[C(t-1)]_i$ corresponds to active label selection based on the predictive uncertainty, which is likely to choose an outlier. The (reciprocal of the) denominator represents how well the point $X_i$ is \emph{connected} to the unlabeled data points~\cite{AmeFogKap11}. Accordingly, normalizing the uncertainty by this value rules out the possibility of choosing the outlier. However, $\calI'(X_i)$ is based on how much information the unlabeled data points contains about $X_i$, which does not directly represent our final goal: our goal is to choose a data point which minimizes the \emph{overall} uncertainty on all data points---this is directly addressed by $\calI$. 

\paragraph{Zhu~\etal comparison.}
Zhu~\etal~\cite{ZhuLafGha03} proposed an active learning framework that selects the labels by minimizing an estimate of the expected risk. This framework relies on evaluating the prediction accuracy of the updated system for each hypothesized output values. Therefore, the application of this approach is limited to the case in which the output space is finite, \eg, $\{-1,1\}$ in the classification problem. Our approach does not have this limit.

\subsection{Evaluation}
\label{sec:experiments}

\begin{figure}[t]
\centering
\includegraphics[width=0.49\linewidth]{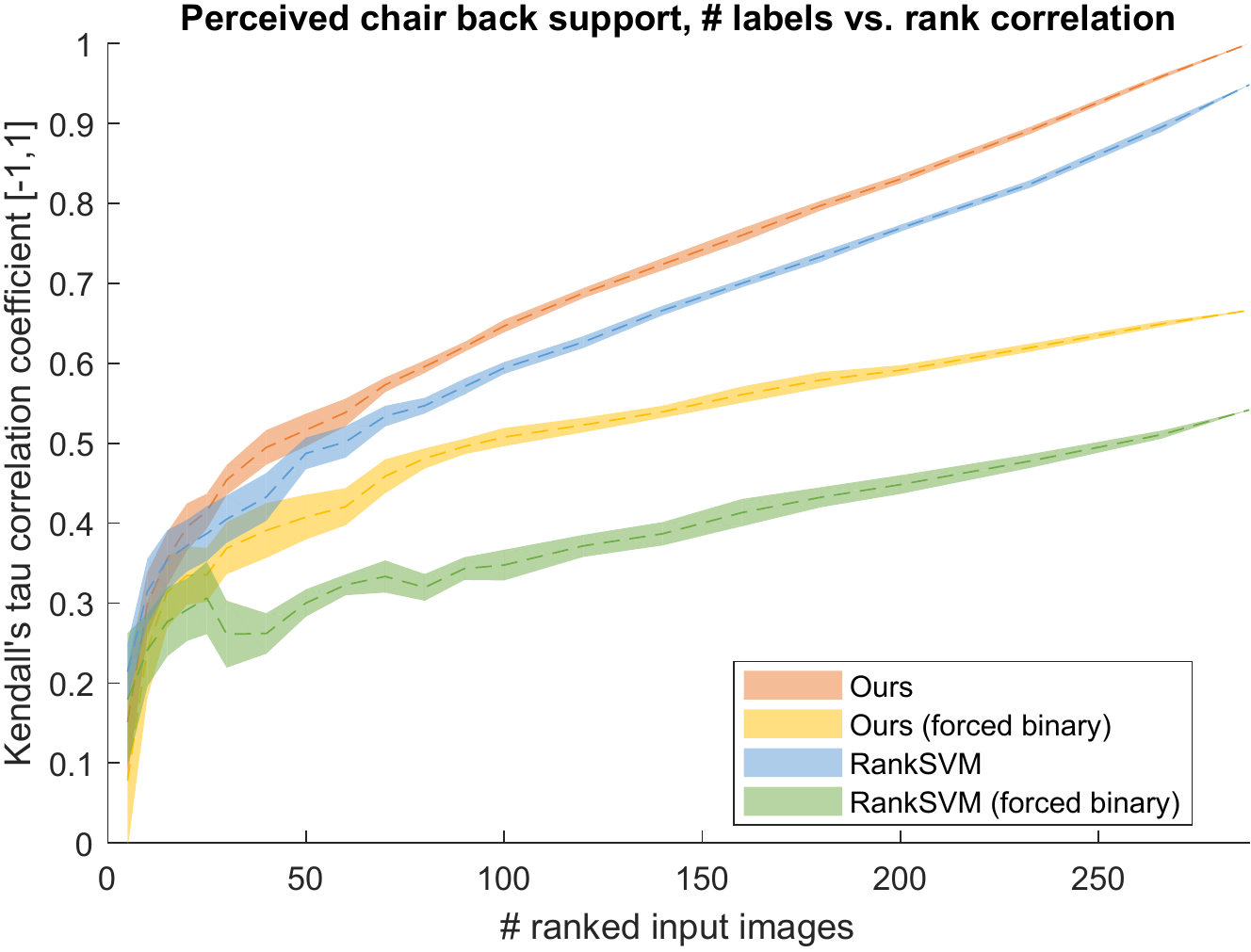}
\includegraphics[width=0.49\linewidth]{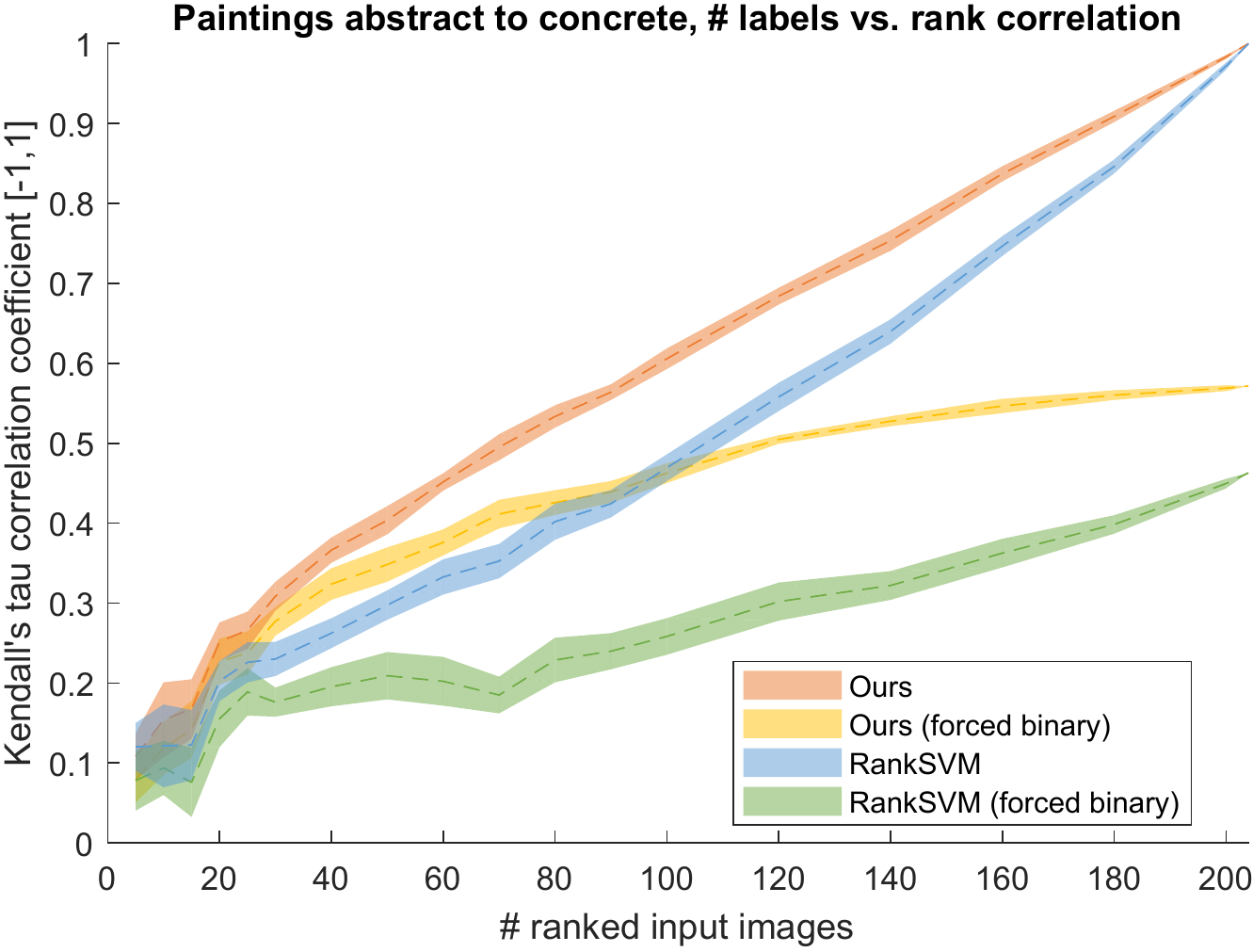}
\includegraphics[width=0.49\linewidth]{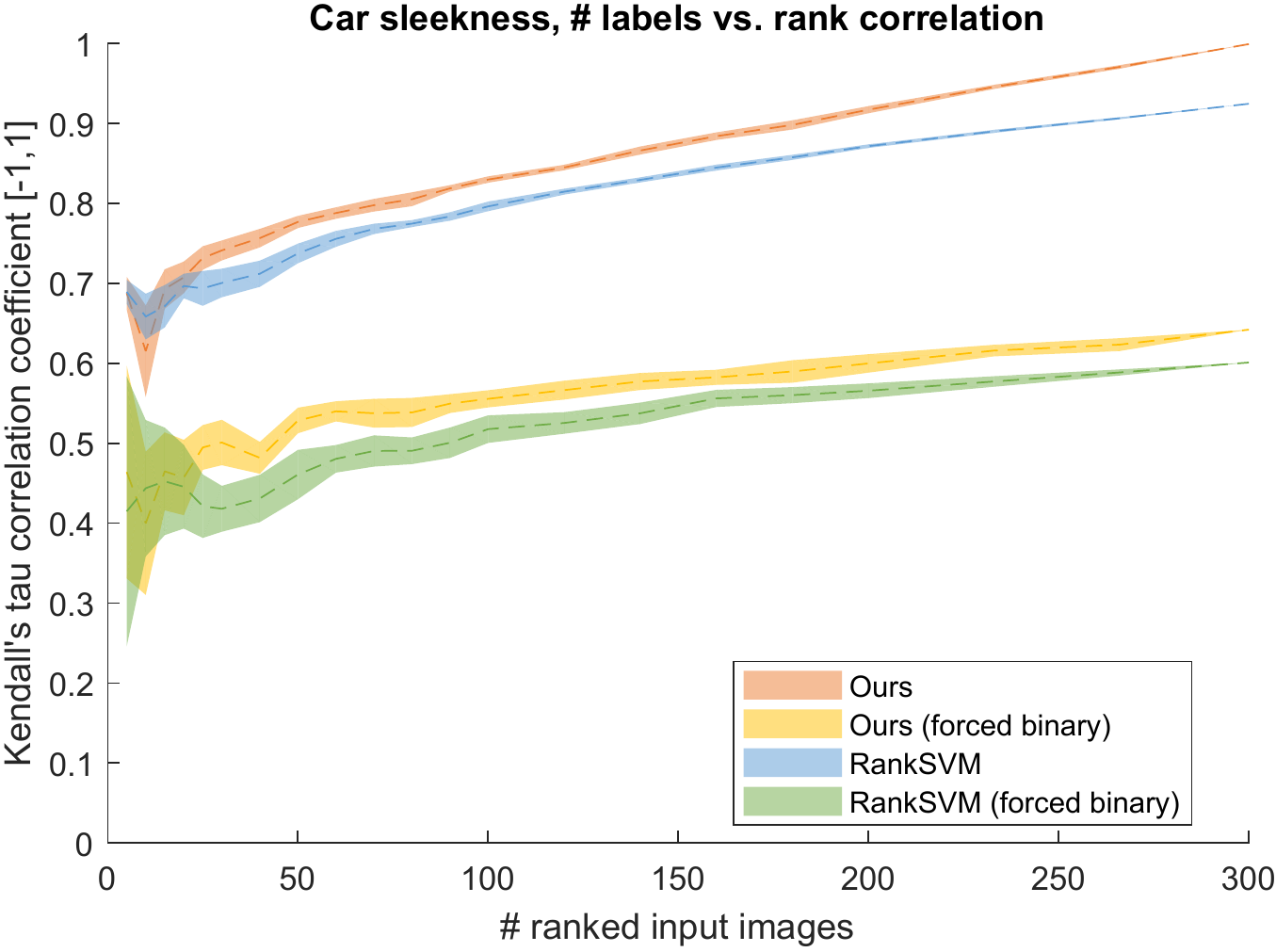}
\vspace{-0.4cm}
\caption{1D criteria learning with varying numbers of labels over 10 random trials (dotted line is the mean, with faded region showing standard error set at 95\% confidence level). For fairness, each technique underwent hyper-parameter optimization to achieve the best result. Our approach is more successful once the number of labels rises above $\approx 20$. \emph{Top left:} Criteria `perceived chair back support' on 288 chair geometries (Sec.~\ref{s:databases}, Fig.~\ref{fig:objects}). \emph{Top right:} Criteria `abstract to concrete' on 201 painting images (Sec.~\ref{s:databases}, Fig.~\ref{fig:paint}). \emph{Bottom left:} Criteria `sleekness' on 300 car geometries (Sec.~\ref{s:databases}, Fig.~\ref{fig:cars})}
\label{fig:performance}
\end{figure}

\paragraph{1D criteria learning.}
\label{s:qualitative}

We evaluate our interactive LG adaptation on user rank data versus two techniques: discrete-only RankSVM approaches~\cite{ParGra11,ShenLin2013}, and a `forced binary' version of our approach (and RankSVM) where labels are set either to $1$ or $-1$ to simulate a simpler yes/no interaction. We compare over increasing numbers of randomly chosen labels, with the remaining data points used as unlabeled examples (10 trials, averaged). Our approach improves performance over both baselines once the number of labels surpasses 20 (Fig.~\ref{fig:performance}).\footnote{For quantitative evaluation of the batch LG regularizer on objective measures, please see Kim~\etal~\cite{KimTompkin2015}.}

\paragraph{2D criteria learning.}

\begin{figure}[t]
	\centering
	\includegraphics[width=\linewidth]{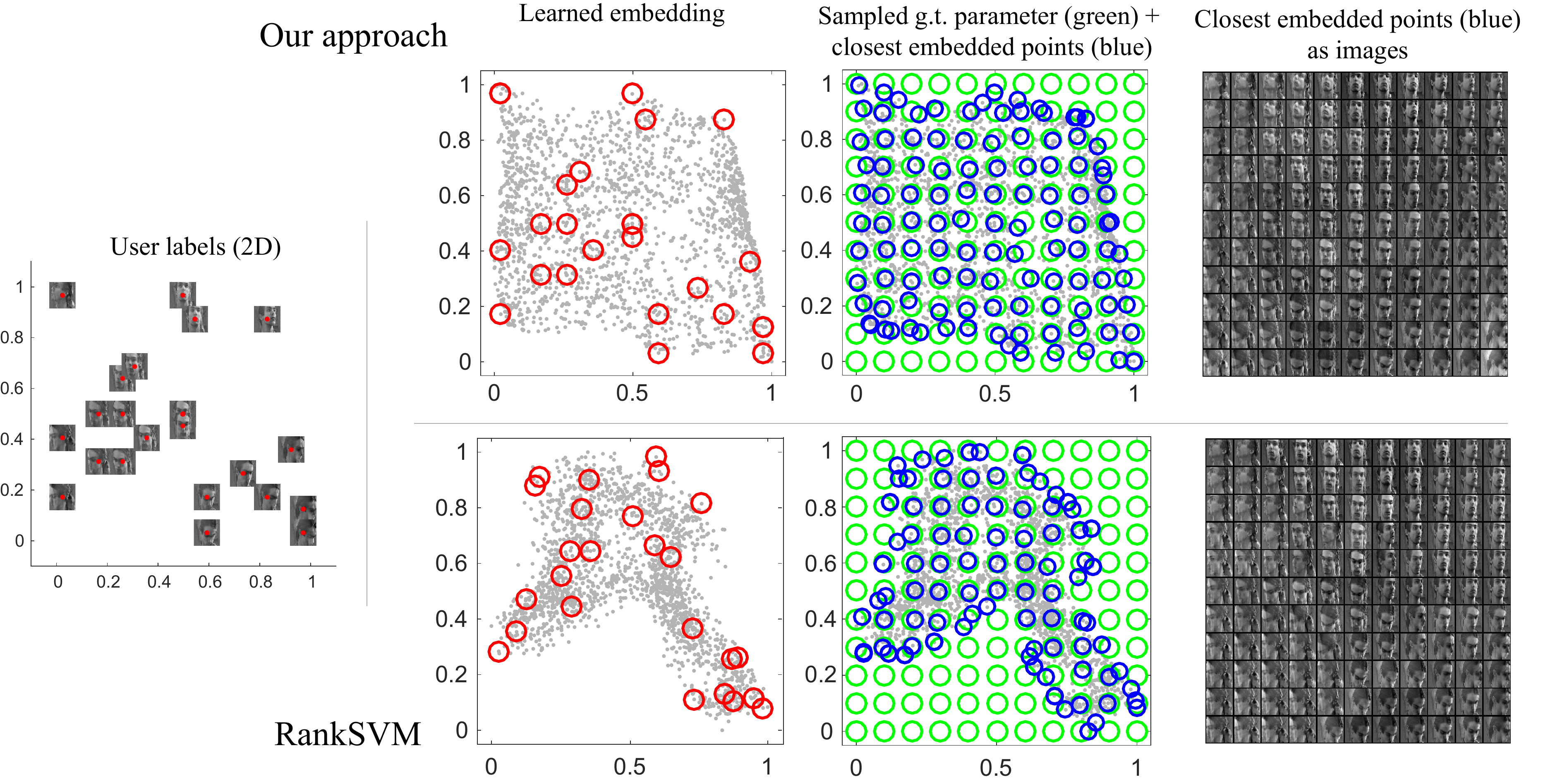}
	\caption{A user labels 25 face images by rotation angles (far left) to learn a criteria over $2,000$ images. \emph{Top:} Our approach. \emph{Bottom:} RankSVM. \emph{Left to right:} 1) learned embedding, showing labeled points as red circles, with our approach better maintaining the spatial layout. 2) Sampling ground truth labeled points uniformly (green circles) to see where their embedded points lie (blue circles), with our approach better reproducing the original uniform sampling. 3) Corresponding images to the blue circles in 2). Please zoom for detail.}
\label{fig:facepose}
\end{figure}

For expert users, we can relax the ranking interface convenience and allow 2D criteria via direct positioning. Training labels are positioned in a 2D unit domain, effectively defining a geometric slider space. The labels are 25 faces selected randomly from 2,000 face images of a single person~\cite{VerVla06}, describing horizontal and vertical rotations (Fig.~\ref{fig:facepose}). Objectively assigning pose angle is difficult, and so the labels are perceptual approximations. Compared to RankSVM, our embedding better reproduces the user's intention with a more even spread over the output space. Since RankSVM does not enable users to define a `geometry' in parameter space, coordinating more than one parameter in this way can be challenging.

\paragraph{Active label suggestion.} 
\label{sec:expactivelearning}

\begin{figure}[t]
\centering
\fcapside[\FBwidth] 
{\caption{Active label selection performance (automatically picking the top label suggestion) as mean absolute error vs. ground truth for learning \emph{body weight} as a criteria on the CAESAR database. \# labels is subsamped by 2 for display. Error bar lengths correspond to twice the standard deviations (std.); please note the smaller stds.~of our proposed full and sparse methods. }
\label{f:als}}
{\includegraphics[width=\linewidth]{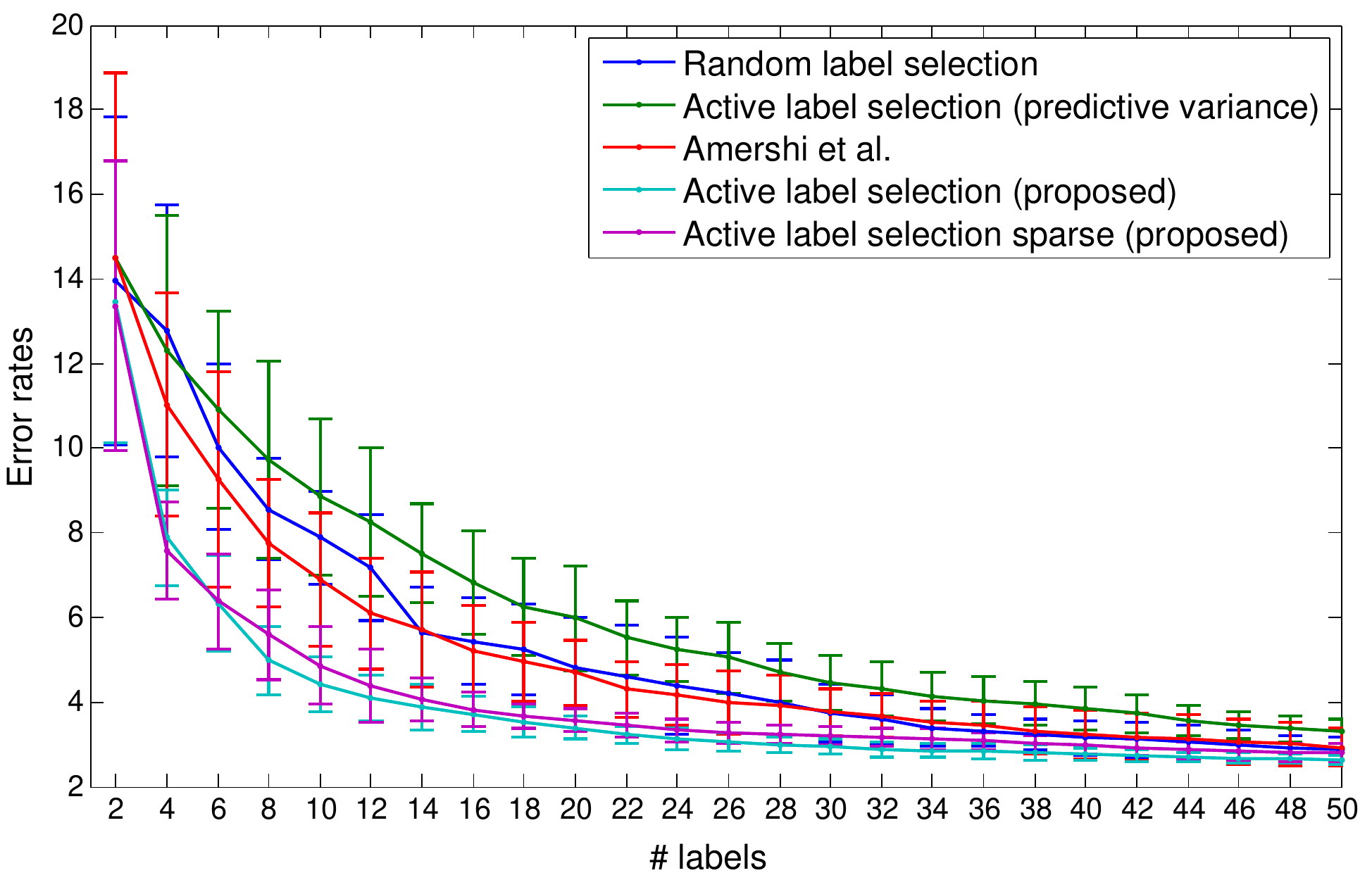}}
\end{figure}

We compare our performance to 1) random label selection, 2) predictive uncertainty, and 3) Amershi~\etal~\cite{AmeFogKap11} adapted to our semi-supervised setting (Fig.~\ref{f:als}). Over 10 trials on the CAESAR dataset (Sec.~\ref{s:databases}), we randomly selected a set $\calA$ of $2,000$ data points, with two intial labels $\calB\subset \calA$ and train on $\calB$. We calculate information gain $\calI$ (Eq.~\ref{e:infogain}) for each data point in $\calA\backslash\calB$, assign the best data point a label, and iterate until $|\calB| = 50$. 

Predictive variance only resulted in suggesting outliers which led to worse results than random selection. The adapted algorithm of Amershi~\etal~improves upon random selection, while our new algorithm shows further improvement, especially when the number of labels is low. The computational complexity of Amershi~\etal~and ours are equal. We also demonstrates that our proposed sparse eigen-decomposition-based approximation (\emph{Active label selection sparse (proposed)}) produces comparable results to the full information-gain-based method (Eqs.~\ref{e:infogain} and \ref{e:rankoneupdate}). We used only the first 100 ($r=100$) eigenvectors. Calculating the selected eigenvectors of a sparse matrix does not require explicitly generating a dense matrix of size $u\times u$, as the eigen-decomposition can be efficiently updated after an initial pre-interaction computation (Eq.~\ref{e:eigenupdate}).

\paragraph{Computation complexity.}
This depends on the number of data points $u$, the number of nearest neighbors $k$ in building the LG regularizer~\cite{KimTompkin2015}, the rank of the sparse approximation $r$ (Eq.~\ref{e:approx}), and the number of non-zeros entries in the resulting regularization matrix $H$. This lies in-between $O(uk)$ and $O(uk^2)$, depending on the well-behavedness of neighborhoods (where $O(uk^2)$ corresponds to random neighbors). $H$ is built once per dataset as a preprocess. The dimensionality of the data model affects only the construction of the regularizer. At interaction time, complexity depends only on the number of data points. For the incremental step, we randomly select $1,000$ candidate data points among $u$ points, and choose the one which maximizes the information gain (Eq.~\ref{e:updateequ}). 

On CAESAR, with $u=4,258$ and $k=20$, the preprocess takes 14 seconds. Solving the system took $0.5$ seconds for standard batch \emph{LG} approach. For active label suggestion, in non-incremental batch LG learning, estimating the predictive variance for each of the $4,258$ data points requires re-training the entire system per data point, which takes $\approx 35$ minutes. In contrast, our algorithm enables suggesting the best data point across all examples in $1.5$ seconds. All timings were on an Intel Xeon 3GHz CPU in MATLAB.

For a larger $60,000$ item database\footnote{We use MNIST purely for its size, though it would be unlikely to use our system on this database.} and $k=10$, the preprocess takes $\approx 10$ minutes, with $5$ seconds per solve. A database of this size requires our sparse eigen-decomposition-based approach to suggest labels, and this takes about $2$ seconds per label. Generally, only a handful of active labels are needed for suggestion, \eg, top 5, and so this still allows (if slower) interaction. Naively calculating the information gain (Eq.~\ref{e:infogain}) took around $4$ hours, while (dense) incremental selection (Eq.~\ref{e:rankoneupdate}) is infeasible due to the prohibitively large memory requirement.

\paragraph{Parameters.}
As default parameters, we set regularization weight $\lambda=10^{-6}$, nearest neighbors $k=20$, and estimated dimensionality of the underlying high-dimensional manifold $m$ in feature vector space varies between 5--20 with database size. Parameters can vary per database. While a hyper-parameter search against a test set is at odds with many motivating applications, \eg, initially assessing a just-collected database, in some applications these parameters could be set by a `database curator', after which end users would define their own criteria. 

\begin{figure}[t]
    \centering
    \fcapside[\FBwidth]
    {\includegraphics[width=9cm]{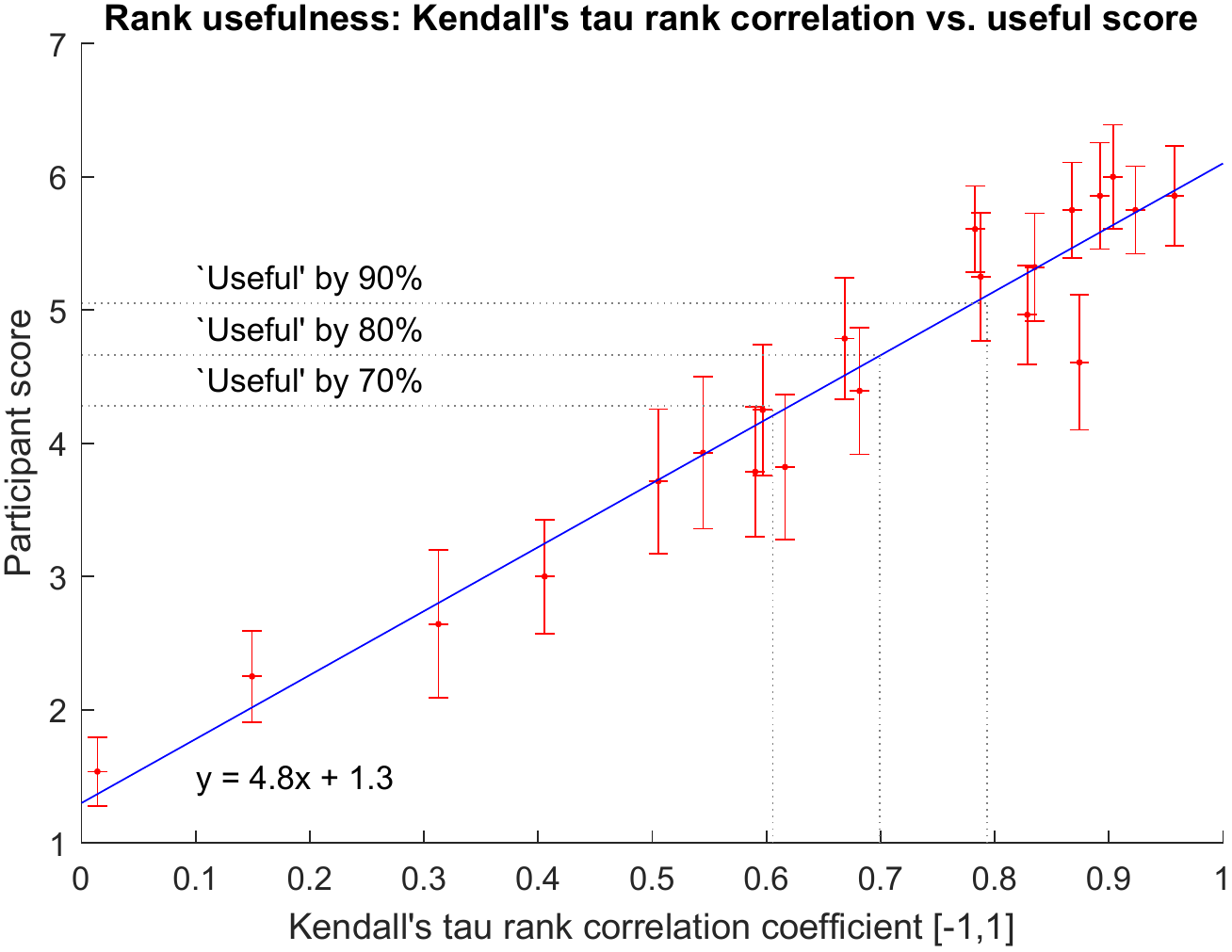}}
    {\caption{Ordering agreement plotted against participant score. At 90\% acceptance, $9\times$ more pair orderings must agree than disagree; at 80\%, $5.9\times$ more; at 70\%, $4.3\times$ more.}\label{fig:perceptualusefulness}}
\end{figure}

\paragraph{Rank usefulness.}
To probe the performance required for useful orderings, we asked 28 users to rate candidate orderings of 50 items against a hypothetical ideal ordering. Users assigned to each candidate rank a 7-point Likert scale score and a binary `useful/not' classification. Candidate orderings were real results generated by our system, which spanned the space between random selection and ground truth. The ideal orderings were visible for comparison. First, the experiment was thoroughly explained and users completed a training example. Then, users ranked 24 candidate rankings in a random order, taken in equal portion from the three `Objects' geometry databases.

We compare the Kendall's tau rank correlation coefficient of the different candidate rank orderings with the average Likert score attributed by the participants (Fig.~\ref{fig:perceptualusefulness}). Assuming our scale data to be approximately continuous, we fit a line to describe the relationship of usefulness ($y$) to KT ($x$) as $y=4.8x + 1.3$. This allows us to approximate the number of labels required for useful criteria. We compute $\tau$ thresholds for 70\%, 80\%, and 90\% acceptance rates. For 70\% of participants to be satisfied, $\tau=0.61$; at this level, $4.3\times$ more pairwise orderings must agree than disagree. For 80\% of participants to be satisfied, $\tau=0.70$, and $5.9\times$ more pairs must agree than disagree. For 90\% of participants to be satisfied, $\tau=0.80$, and $9\times$ more pairs must agree than disagree.

\paragraph{Human variation.}
Some criteria are inherently ambiguous. In one application, we organize paintings along the criteria ``abstract to concrete''. This task has no well-defined answer: while there is a general trend that can be agreed by most people, the specific rank positions of the paintings are open to interpretation. As such, one measure for the performance that our system should aim to achieve to be useful is the level of agreement between humans.

\begin{figure}[t]
    \centering
    \fcapside[\FBwidth]
		{\includegraphics[width=9cm]{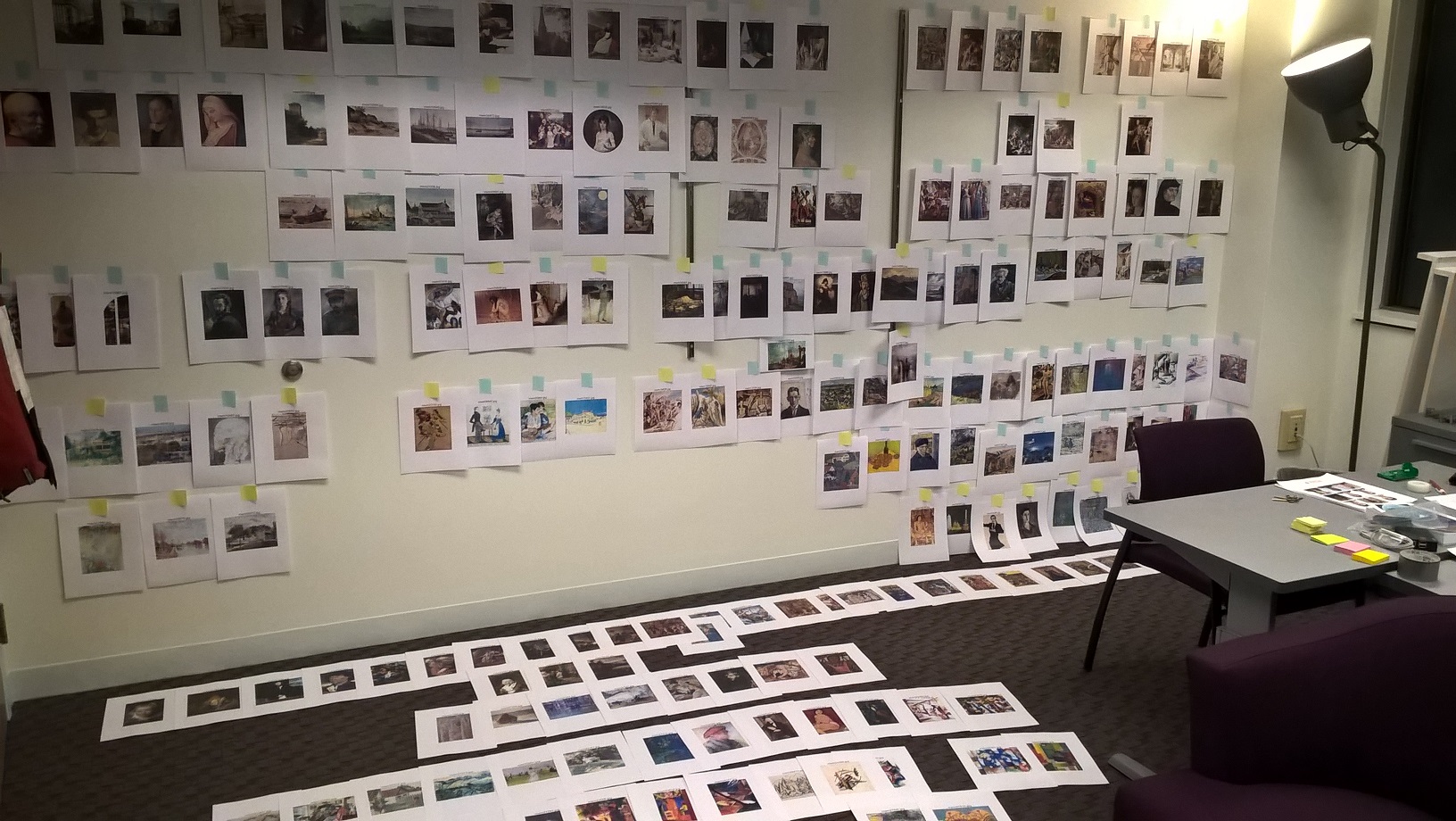}}
    {\caption{Ranking 201 paintings by printing each on paper and arranging them over a large wall (and floor) workspace.}\label{fig:paintingwallranking}}
\end{figure}

To investigate the human variance of ranking paintings, we asked seven users to rank 201 images along the criteria ``abstract to concrete''. The users were aged 20--40, 2 female and 5 male, with varying artistic experience from regular painters to casual knowledge. During this `ground truth' generation task, many visual comparisons must be made across the rank. Completing this task on a computer is currently difficult because displays are too small to show all images at once with sufficient size for comparison. Thus, we printed each image on paper and asked participants to tape them to a very large wall. The wall acted as the `rank space', and participants were free to use it as they wished. Most used some combination of bubble, insertion, and merge sorts (Fig.~\ref{fig:paintingwallranking}).

We compute Kendall's tau rank correlation coefficient between each pair of user rankings, producing 21 coefficients (Tab.~\ref{tab:kunstuservariation}). The mean coefficient is (coincidentally also) 0.619, with standard deviation of 0.043, and range 0.530--0.684. There is only moderate agreement between participants. This shows the difficulty of the task and the need for an interactive ranking system which can more easily create personal criteria. However, it also provides us evidence, at least for this task, that a system should aim to achieve this level of performance to reach the `agreement' among participants. To achieve approximately the `subjective average' performance across participants in this task with our interface would require approximately 100 labels (Fig.~\ref{fig:performance}, top right). 

On average, participants spent 158 minutes to produce their rank---ordering a complex database can take time! We initially developed a mouse-based list drag and drop computer interface for this task, with quick buttons to move items to different sections of the 201-item list as a quick approximate pre-sort (e.g., `Move to start/10\%/20\%\ldots /end'). However, this was slower than using paper and a large physical rank space (Fig.~\ref{fig:paintingwallranking}). 

\newfloatcommand{tcapside}{table}[\capbeside]

\begin{table}[b]
    \centering
    \tcapside[\FBwidth]
    {
    \begin{tabular}{llllllll}  
        \toprule
        Par. & 1 & 2 & 3 & 4 & 5 & 6 & 7 \\
        \midrule
        1 & - &   0.6296 &   0.6067 &    0.5269 &    0.6094 &    0.6028 &    0.6032 \\
        2 & 0.6296 &   - &   0.6806 &    0.6443 &    0.6839 &    0.6667 &    0.6130 \\
        3 & 0.6067 &   0.6806 &   - &    0.5760 &    0.6449 &    0.6237 &    0.5941 \\
        4 & 0.5269 &   0.6443 &   0.5760 &    - &    0.6444 &    0.6401 &    0.5342 \\
        5 & 0.6094 &   0.6839 &   0.6449 &    0.6444 &    - &    0.6586 &    0.5661 \\
        6 & 0.6028 &   0.6667 &   0.6237 &    0.6401 &    0.6586 &    - &    0.6407 \\ 
        7 & 0.6032 &   0.6130 &   0.5941 &    0.5342 &    0.5661 &    0.6407 & - \\
        \bottomrule
    \end{tabular}
    }
    {\caption{Kendall's tau rank correlation coefficients computer pairwise between seven participants, each asked to rank 201 paintings by the subjective criteria `abstract to concrete'. There is mild agreement across all participants, showing both the difficulty of the task and a baseline level of performance required to satisfy the group on average.}\label{tab:kunstuservariation}}
\end{table}

\subsection{Interactive prototype}
\label{s:prototype}

\begin{figure}[t]
\centering
\includegraphics[width=0.85\linewidth]{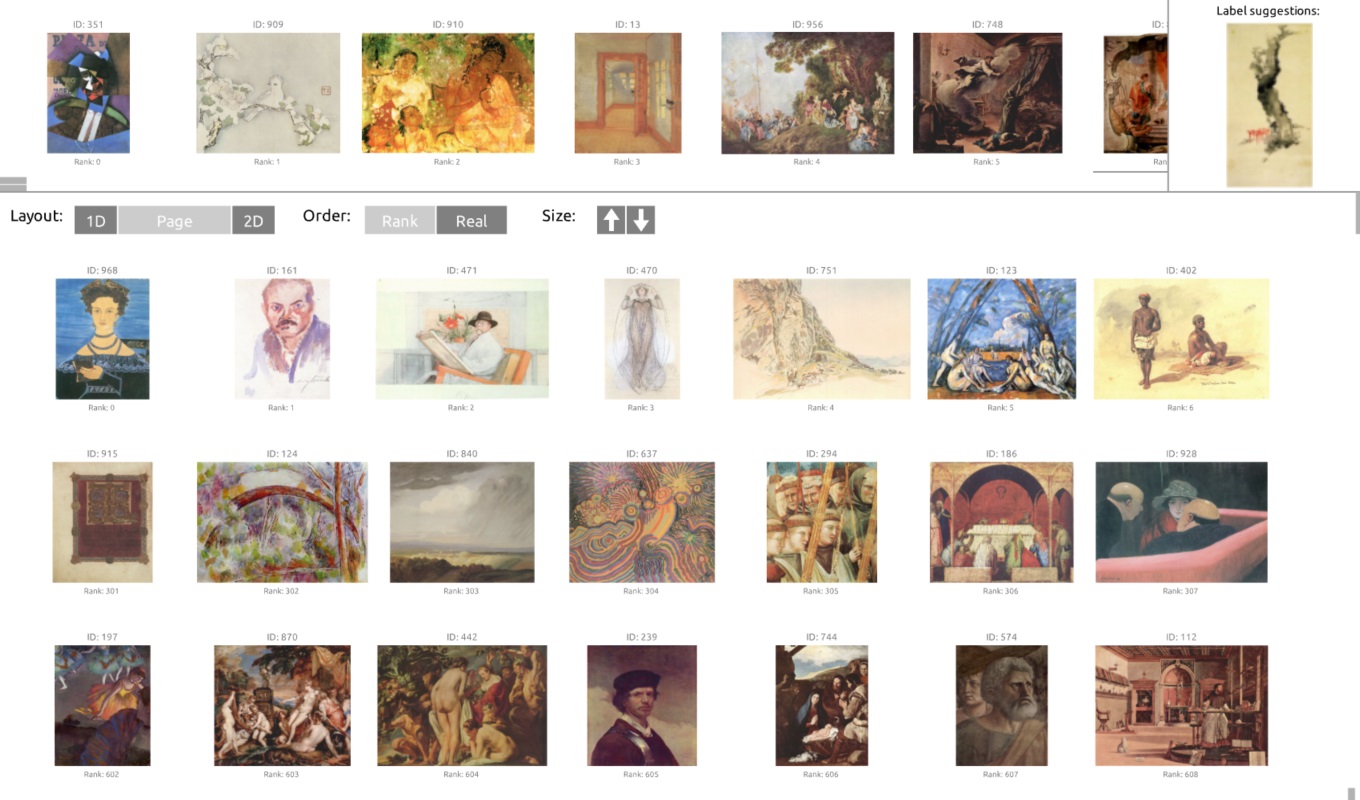}
\caption{Our prototype interface to define criteria on visual databases, in this case of paintings. The user is attempting to describe a criteria which spans `abstract to concrete'. \emph{Top:} The user ranking of examples. To the far right is the suggested next item to label. \emph{Bottom:} The database, ordered by the criteria, displayed in western page order. To see the broad trend across all items, the user has decided to show only a subset of the data sampled across all items. We can see that drawings and pictographic representations are at the top left, and more realistic or concrete depictions are at the bottom right.}
\label{fig:interface}
\end{figure}

Our prototype interactive interface allows users to describe criteria by ranking examples, receiving active label suggestions, and viewing and manipulating the recovered sliders (Fig.~\ref{fig:interface}). The database is shown to the south, while the user rank is shown to the north. Initially, the database is shown in a random order. The user can drag suggested items to the rank, or drag any database item from the south up to the rank. The items are not `removed' from the database on dragging; merely, this is the interaction metaphor for labeling an example. Every time an item is inserted into the rank or whenever the rank is reordered, we update the criteria and order all items in the database display to the south. Efficient learning is key to this interactivity.

\begin{figure}[t]
\centering
\fcapside[\FBwidth] 
{\includegraphics[width=8cm]{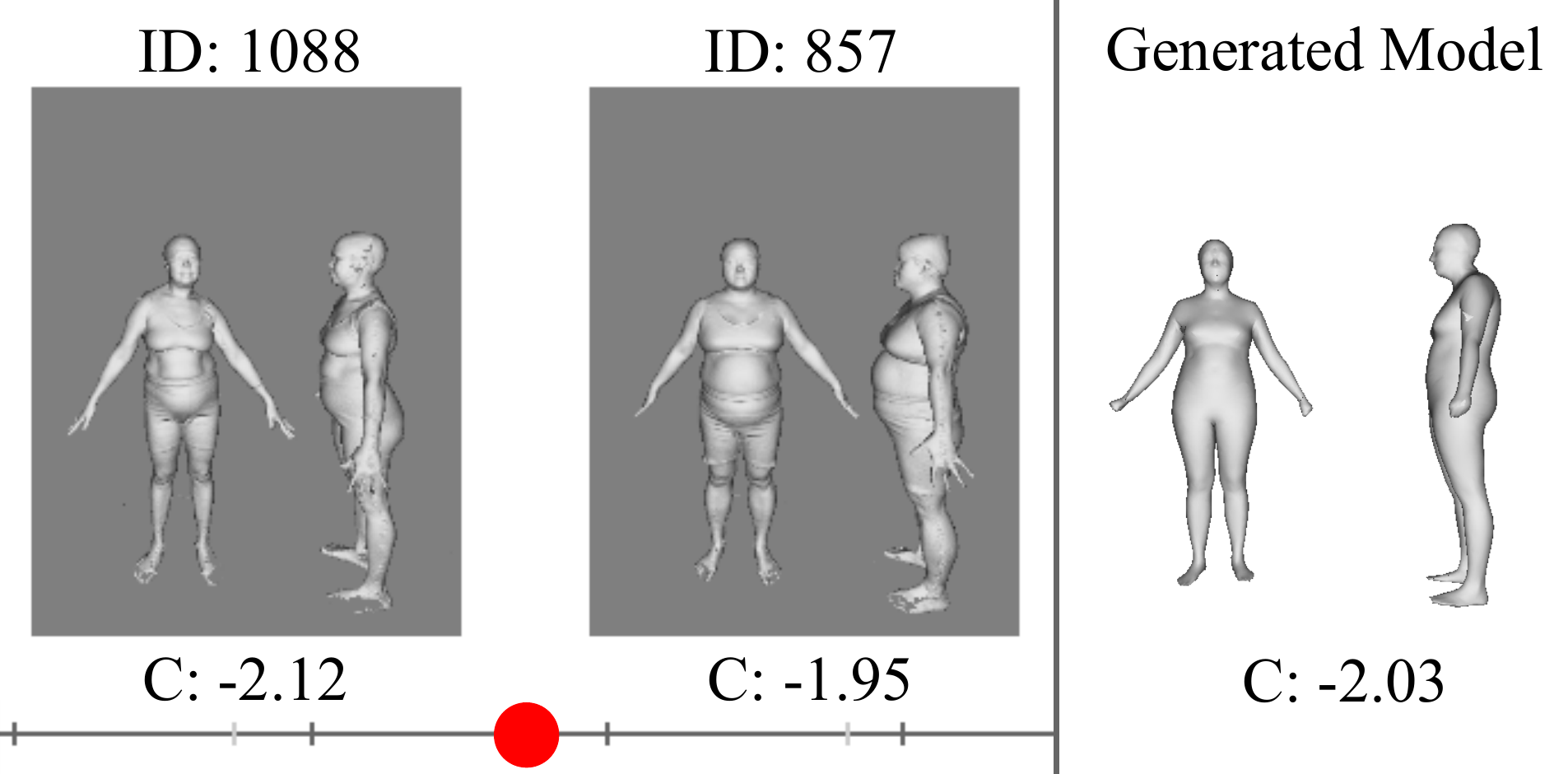}}
{\caption{With an underlying high-dimensional generative model, we can create new examples by interpolating the discovered continuous criteria. For example, given the CAESAR database, we can generate a human form with a shape in between two examples on our criteria.}\label{fig:contentcreation}}
\end{figure}

Criteria may be defined by ranking in 1D. These can be displayed as a linear rank or in western page order, similar to a Web image search where items are laid out by relevance. As our underlying representation is continuous, the criteria can also be displayed as a continuous scale where items are separated by their estimated distances along the scale. For databases which use underlying generative models as their high-dimensional representation, this allows new examples to be generated at specific points along the scale, \eg, at a specific point between two desirable examples (Fig.\ref{fig:contentcreation}).

\begin{figure}[t]
\centering
\includegraphics[width=0.8\linewidth]{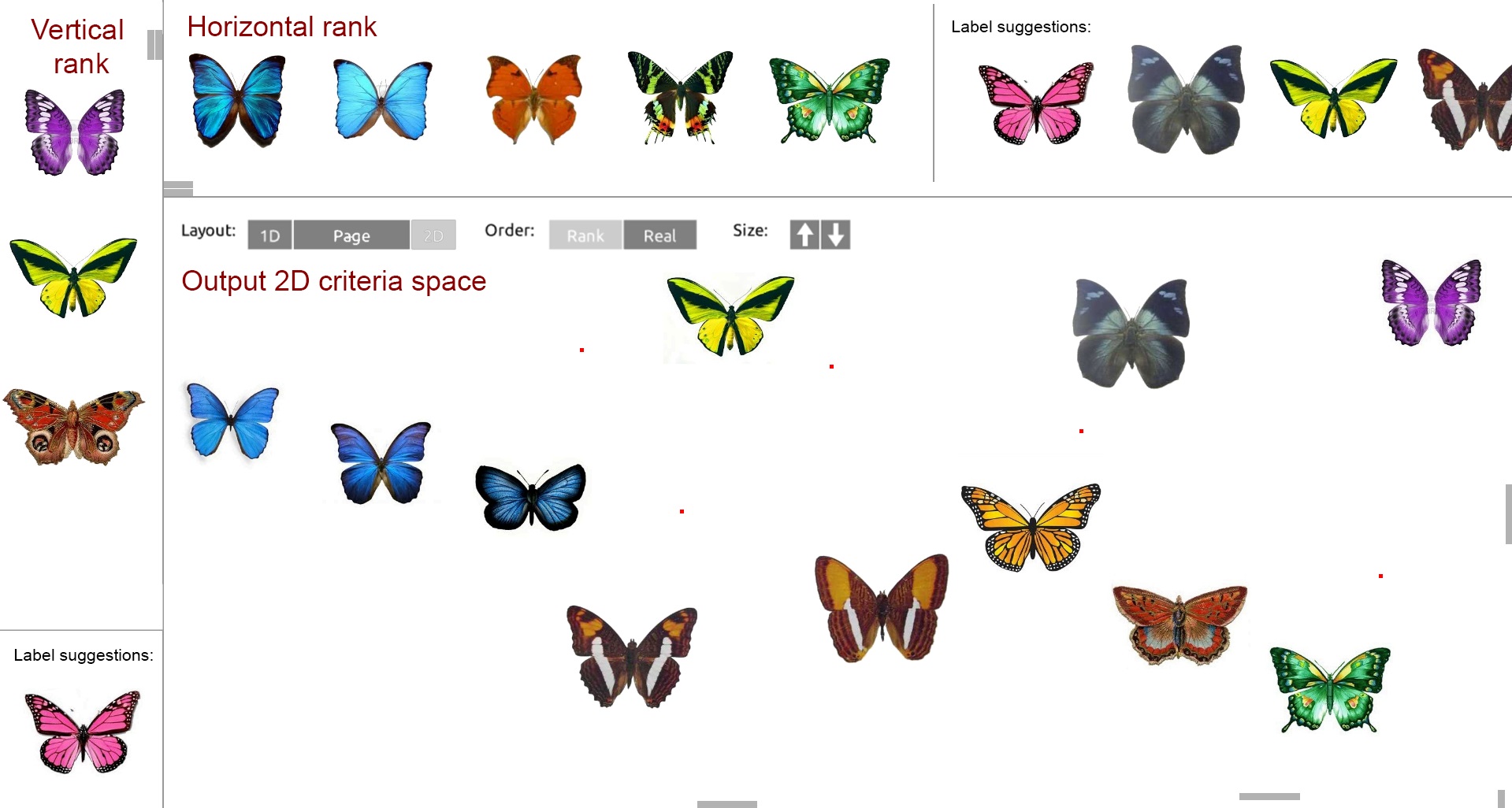}
\caption{Our interface is used to define an output pseudo-2D criteria space from two independent 1D ranks. The horizontal criteria labels are ranked in the bar at the top left (\emph{Label rank 1}), with label suggestions to the right. The vertical criteria is on the far left (\emph{Label rank 2}). In this example, the user has zoomed in to a subregion of the learned criteria space, where representative data points are visible as images and red dots represent the remaining data points.}
\label{fig:interface2}
\end{figure}

Criteria may also be defined in pseudo-2D by ranking with two independent 1D rankings (Fig.~\ref{fig:interface2}). The criteria output is then viewed as a 2D scatter plot, either as a matrix-style rank ordering or as a continuous 2D space from the underlying continuous representation. For experts, we can relax the rank interaction convenience and allow criteria to be defined directly by placing examples on the continuous scale, e.g., Figure \ref{fig:facepose}.

Our databases can be large with many thousands of items, and so we allow the user to reduce the number of presented items to show the broader trend. When viewing the database organized by rank, we allow the user to control a subsampling of the database items. When viewing the database organized along continuous scales, we allow the user to zoom out. Users can also jump quickly to a relevant point in the output criteria by right-clicking on the rank, and vice versa.

\section{Datasets and example applications}
\label{s:databases}
\begin{figure}[t]
    \centering
		\includegraphics[width=\linewidth]{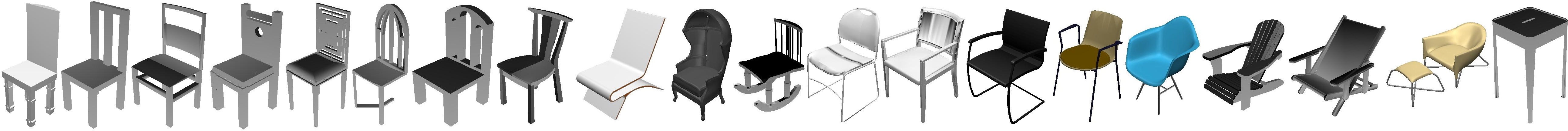}
		\includegraphics[width=\linewidth]{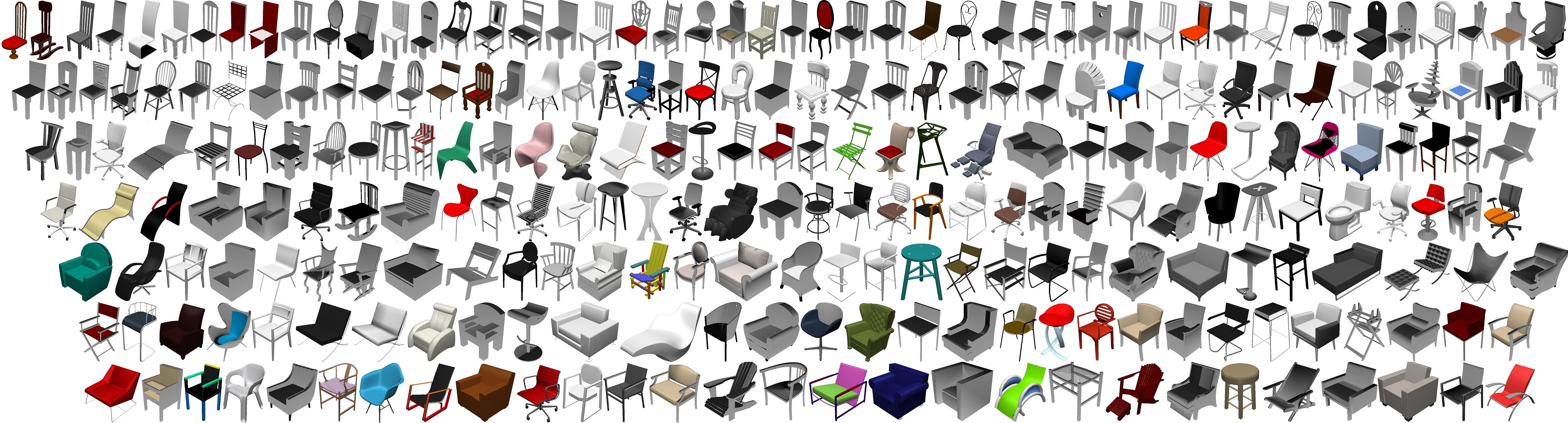}
    \caption{Result at 20 labels. \emph{Top:} Rank preference ($\approx$2 minutes). \emph{Bottom:} Rank propagated to 288 data points (raster order). While there are occasional outliers, we maintain the trend of chairs with high backs, then chairs with arms, and then lounge chairs, in this highly variable database. Figure \ref{fig:performance} evaluates performance across varying numbers of labels.}
    \label{fig:objects}
\end{figure}

\begin{figure}[t]
		\includegraphics[width=\linewidth]{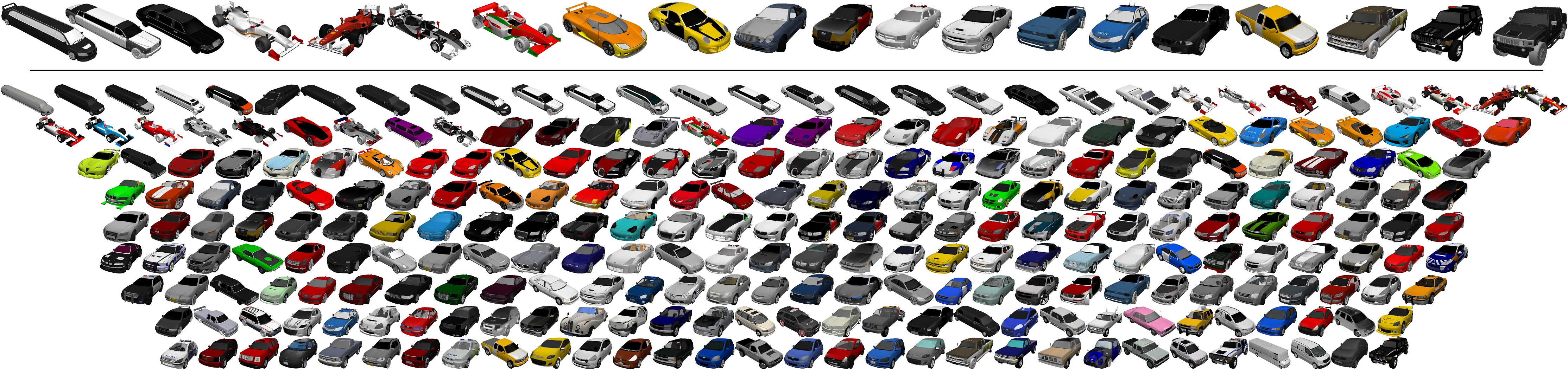}
    \caption{300 items from the cars 3D object database, organized by user criteria `sleekness', via our approach, with the 20 labels provided at the top. Figure \ref{fig:performance} evaluates performance across varying numbers of labels.}
    \label{fig:cars}
\end{figure}

\paragraph{Object geometry.} We use data from Hua~\etal~\cite{HuaSuGui13}: 5,000 chair (Fig.~\ref{fig:objects}), 3,000 airplane (Fig.~\ref{fig:teaser}), and 1,700 car geometries (Fig.~\ref{fig:cars}) are downloaded from Trimble 3D Warehouse, then oriented and scaled to align object features, then locally deformed to better align shared elements, \eg, seat heights for chairs. Once normalized, we voxelize these spaces and extract, for each voxel, the shortest distance to the nearest mesh point to create a distance field which captures the shape variation of each example. Figure~\ref{fig:objects} shows an example for chair style preference. \emph{Related applications}: Shape exploration~\cite{HuaSuGui13} and synthesis~\cite{Maks11} scenarios, especially for large Web collections. 

\begin{figure}[t]
\centering
\includegraphics[width=\linewidth]{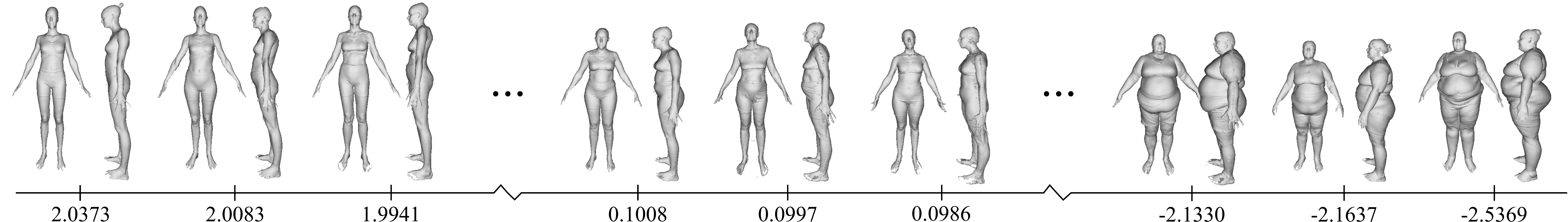}
\caption{2226 CAESAR female body scans are ordered by \emph{height with priority, then girth}, expressed with 50 labels. While some local variation remains, we can identify the general trend.}
\label{figa:caesar}
\end{figure}

\paragraph{Human geometry.} The CAESAR database contains 4,258 human 3D scans, along with ground-truth caliper body measurements. We fit a statistical model to each of the scans \cite{pishchulin15arxiv}, which describes body variations in an abstract 20-dimensional linear shape basis. Fig.~\ref{figa:caesar} shows 2226 female body scans ordered by `height with priority, then girth', which project two orthogonal but related physical attributes down to a single criteria (Fig.~\ref{figa:caesar}).
\emph{Related applications}: Finding useful criteria for statistical shape models instead of semantically-meaningless axes found by dimensionality reduction, \eg, body~\cite{Hasler2009}, face~\cite{Vlasic2005}, or cloth ~\cite{DRAPE2012}.

\begin{figure}[t]
	\centering
	\includegraphics[width=\linewidth]{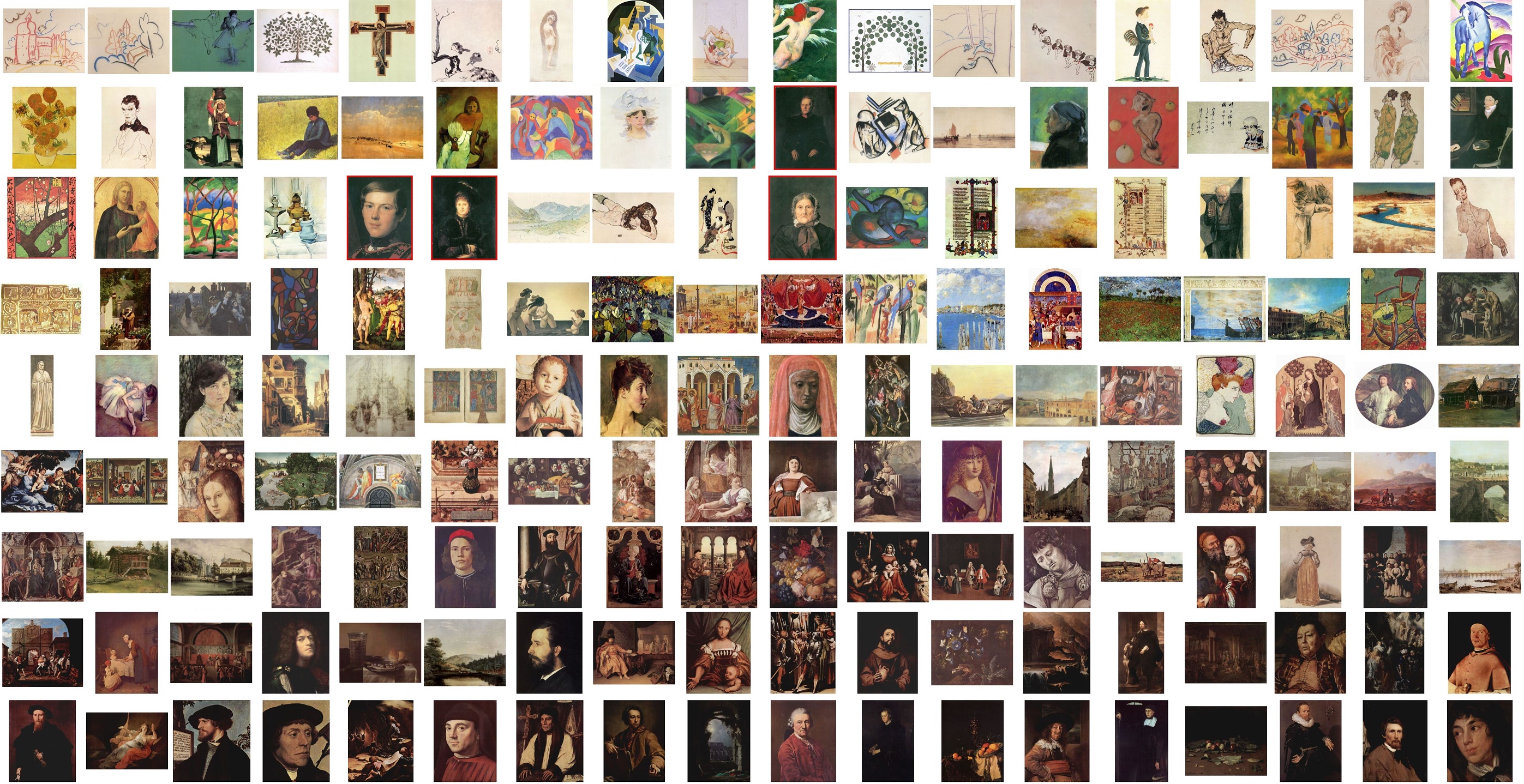}
	\caption{A 162-point subset of 19,808 paintings ordered by the criteria ``abstract to concrete'', sampled evenly along the output interval $[-1.2,1.2]$. Raster order. The images generally increase in realism, with outliers marked in red. While the trend appears dominated by brightness, upon closer inspection (please zoom) the detail is apparent: darker abstract portraits are nearer the top, and brighter concrete landscapes nearer the bottom. Figure \ref{fig:performance} evaluates performance across varying numbers of labels.}
    \label{fig:paint}
\end{figure}

\paragraph{Painting images.} We collected 19,808 paintings covering periods until 1930 (\url{www.zeno.org/kunst}). As a feature vector, we follow Gatys~\etal~\cite{Gatys_2016_CVPR} and use the pre-trained VGG 19-layer convolutional neural network to estimate a set of style matrices, based on computing Gram matrices from the neural response in the first five layers. Figure \ref{fig:paint} depicts a subset of the database after 201 labels were provided for the criteria ``abstract to concrete''. \emph{Related applications}: Arranging cultural heritage or historical artifacts by style/genre~\cite{Cul11,WalFleCun09}.

\paragraph{Face images.} We fit an active appearance model to the Labeled Face Parts in the Wild image database \cite{fiducials_cvpr2011,tzimiro_ICCV2013}. We extract shape and appearance features as the coefficients of their respective PCA spaces. Figure \ref{fig:teaser} shows a result where a user ranked 10 faces by expression, to recover an expression slider covering $417$ males in the database. \emph{Related applications}: Face appearance ranking~\cite{Leyvand2008}, such as personalized attractiveness scales~\cite{Eisentha2006}, or in policing to aid suspect identification by a witness ranking database examples by `more/less like him/her'.

\subsection{Discussion}
\paragraph{Ranking vs. metric distances.}
Interactive ranking makes the intuitive assumption that people find ranking easier than specifying explicit or relative distances between many objects on scales, where the difficulty of this task is proportional to the complexity of the criteria. This ranking choice assumes that scale metric distances depend linearly on the ordering. In principle, this does not incur any loss of generality: Once a criteria scale is generated, we can always non-linearly re-metrize it by performing an order-preserving regression on the final result.

\paragraph{Feature power.}
We focused on designing an interactive semi-supervised learning system suitable for high-dimensional spaces. However, ultimately, the performance of our algorithm also depends on the data features. We use existing feature representations from corresponding related problems (\eg, distance fields for geometric objects); however, in general, the best features depend on the specifics of the database. Here, more sophisticated methods for feature extraction and selection are applicable, \eg, using automatic relevance determination~\cite{RasWill06}.

\paragraph{Non-zero-knowledge vs. exploiting existing labels.}
We intentionally designed our system for the general `zero-knowledge' case where there are no existing labels, \eg, using criteria definition as queries to assess what is in a Web-scraped database. However, in the wild, some labels may exist. Taking inspiration from Chaudhuri~\etal~\cite{ChaKalGig13}, who extend a supervised learning framework \cite{ParGra11}, it may be possible to create an interactive criteria exploration system where crowd-sourced labels `guide' our regularizer. This may pull the solution towards the `average' case, which removes absolute subjectivity, but it may also be possible for this guiding to be parameterized and under user control. Further, we only address single session labeling scenarios, but over multiple sessions it may be possible to apply structured labeling techniques to reinforce previous decisions or allow criteria evolution \cite{KulAmeCar14}.

\section{Conclusion}
\label{s:conclusion}
Interactive database exploration is a difficult problem, and a lack of labels requires efficient analysis via semi-supervised learning. To achieve this, we introduced an incremental version of the LG regularizer, which has superior performance to RankSVM. This enables us to guide the user via a new active label suggestion method, which estimates the information gain of labeling a particular data point. Our approach is sparse to enable LG active learning for large datasets. We show an interactive system which uses rank labels to quickly and easily create criteria sliders, and demonstrate its use across datasets of images and geometry.

\section*{Acknowledgements}
We thank Qi-Xing Huang, Leonid Pishchulin, Thomas Helten, and all of our study participants, particularly Atsunobu Kotani, Frances Chen, Gary Chien, Numair Khan, and Eleanor Tursman. Kwang In Kim thanks EPSRC EP/M00533X/2, and James Tompkin and Hanspeter Pfister thank the DARPA Memex program.

\vfill
\bibliographystyle{ieee}
\bibliography{bib/biblio}
\end{document}